\documentclass[10pt,twocolumn,letterpaper]{article}

\usepackage[pagenumbers]{cvpr} 

\definecolor{cvprblue}{rgb}{0.21,0.49,0.74}
\usepackage[pagebackref,breaklinks,colorlinks,allcolors=cvprblue]{hyperref}
\usepackage{multirow}
\usepackage{fontawesome}
\usepackage{xcolor}
\definecolor{lightred}{HTML}{d63366}
\newcommand{\ModelName}{FireEdit\xspace}
\def\paperID{7916} 
\def\confName{CVPR}
\def\confYear{2025}

\title{\ModelName: Fine-grained Instruction-based Image Editing via Region-aware \\ Vision Language Model}

\author{
Jun Zhou$^{1,2}$\footnotemark[1]~, 
Jiahao Li$^{2}$, 
Zunnan Xu$^3$\footnotemark[1]~,
Hanhui Li$^1$, 
Yiji Cheng$^2$, 
Fa-Ting Hong$^4$\footnotemark[1]~,
\\
Qin Lin$^2$,
Qinglin Lu$^2$, 
Xiaodan Liang$^1$\footnotemark[2]~
\\
$^1$Shenzhen Campus of Sun Yat-sen University~
$^2$Hunyuan, Tencent~
$^3$Tsinghua University~
$^4$HKUST
}

\begin{document}
\renewcommand{\thefootnote}{\fnsymbol{footnote}}
\maketitle

\footnotetext[1]{Work done during the internship at Tencent.} 
\footnotetext[2]{Corresponding author.}

\begin{abstract}
Currently, instruction-based image editing methods have made significant progress by leveraging the powerful cross-modal understanding capabilities of vision language models (VLMs). However, they still face challenges in three key areas: 1) complex scenarios; 2) semantic consistency; and 3) fine-grained editing. To address these issues, we propose FireEdit, an innovative \textbf{F}ine-grained \textbf{I}nstruction-based image editing framework that exploits a \textbf{RE}gion-aware VLM. 
FireEdit is designed to accurately comprehend user instructions and ensure effective control over the editing process. 
Specifically, we enhance the fine-grained visual perception capabilities of the VLM by introducing additional region tokens.
Relying solely on the output of the LLM to guide the diffusion model may lead to suboptimal editing results.
Therefore, we propose a Time-Aware Target Injection module and a Hybrid Visual Cross Attention module. The former dynamically adjusts the guidance strength at various denoising stages by integrating timestep embeddings with the text embeddings. The latter enhances visual details for image editing, thereby preserving semantic consistency between the edited result and the source image. By combining the VLM enhanced with fine-grained region tokens and the time-dependent diffusion model, FireEdit demonstrates significant advantages in comprehending editing instructions and maintaining high semantic consistency. Extensive experiments indicate that our approach surpasses the state-of-the-art instruction-based image editing methods. Our project is available at \href{https://zjgans.github.io/fireedit.github.io/}{\textcolor{lightred}{FireEdit}}.
\end{abstract}    

\begin{figure}[ht]
\centering
\setlength{\abovecaptionskip}{5pt}
\includegraphics[width=\linewidth]{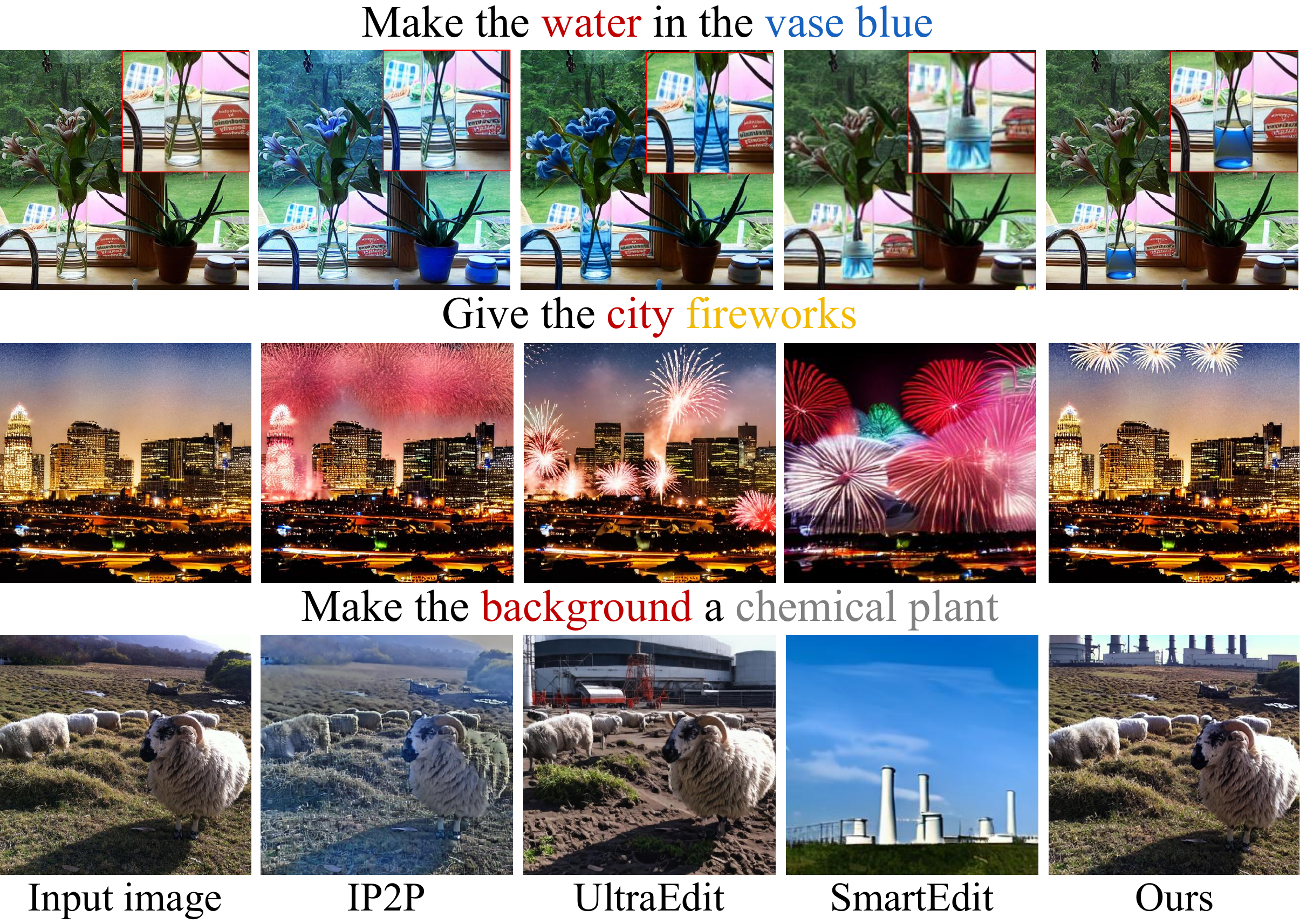}
\caption{
Our framework leverages a vision language model (VLM) to guide instruction-based image editing. 
Our primary innovation is the introduction of region tokens, which enable the VLM to accurately identify edited objects or areas in complex scenarios while preserving high-frequency details in unintended regions during image decoding.}
\label{fig:teaser}
\vspace{-15pt}
\end{figure}

\section{Introduction}
\label{sec:intro}
As a critical area of research in multimedia interaction, image editing has consistently garnered significant attention from both academia and industry. The emergence of diffusion models~\cite{ho2020denoising,ho2021classifier,rombach2022high} and autoregressive models~\cite{sun2024autoregressive,tian2024visual} has reshaped the fields of image generation~\cite{huang2024consistentid,li2024hunyuan,chong2024catvton,li2025blobctrl} and video generation~\cite{jin2024alignment,li2024dispose,xu2024mambatalk,kong2024hunyuanvideo,xu2025hunyuanportrait}. This advancement provides enhanced control and flexibility for generative image editing, enabling the development of numerous commercial products. By transferring the discrepancies between input and target text descriptions to a cross-modal implicit mapping of intermediate latent features, the desired regions of real images can be modified~\cite{hertzprompt,mokady2023null,kawar2023imagic,brack2024ledits++,lin2025mvportrait,lingdrag}. Precise editing is achievable when region masks are available. However, fine-grained text descriptions and region masks are often difficult to obtain.

Instruction-based image editing methods eliminate the reliance on these seemingly essential conditions and are widely adopted in the community for their simplicity, user-friendliness, and efficiency. 
IP2P~\cite{brooks2023instructpix2pix} is a pioneering work in instruction-based editing, allowing users to express their modification intentions for a specified image using straightforward natural language. To achieve precise and high-fidelity image editing, recent methods focus on large-scale generation of high-quality input-target-instruction triples  ~\cite{zhang2024magicbrush,sheynin2024emu,hui2024hq,zhao2024ultraedit,ge2024seed,yu2024anyedit} and exploit the remarkable reasoning and comprehension capabilities of LLMs~\cite{fuguiding,huang2024smartedit,yu2024promptfix}. 
Despite the significant progress made by these methods in advancing instruction-based image editing, several shortcomings persist in the following scenarios: 1) complex instructions; 2) intricate scenes (e.g., multi-object environments and urban streets); and 3) fidelity degradation, which encompasses undesirable changes in style, texture, color, and other visual details. As illustrated in Figure~\ref{fig:teaser}, current methods either demonstrate inaccurate localization or inadvertently alter unintended objects. These challenges arise from two primary factors. First, the CLIP text encoder~\cite{radford2021learning} employed by Stable Diffusion~\cite{rombach2022high} struggles to extract the desired modifications from the instructions provided by general users. In complex scenarios, even when users invest considerable time crafting their instructions, aligning the text with the editing regions in the input images remains a challenge. Second, the data pairs used for training are predominantly synthesized by text-to-image (T2I) generation models~\cite{rombach2022high,hertzprompt,podellsdxl,esser2024scaling}, which inevitably introduce implicit biases~\cite{lee2024holistic,friedrich2023fair}.
Consequently, models trained on ambiguous or imprecise data often struggle to accurately identify the editing object and may lose high-frequency details in unintended areas of the input image.

To address these limitations, we propose FireEdit, a novel end-to-end, instruction-based image editing method. FireEdit primarily comprises a Vision Language Model (VLM) and a diffusion model. The VLM converts user-provided editing instructions into an edit representation that can be comprehended by the generative model. Previous methods~\cite{fuguiding,huang2024smartedit} have employed VLM to improve the model's understanding of editing instructions.
MGIE~\cite{fuguiding} utilizes VLM to generate concise expression instructions, enhancing simple instructions with greater visual imagination. SmartEdit~\cite{huang2024smartedit} addresses complex instructions by implementing a bidirectional interaction module that promotes mutual understanding between visual features and VLM outputs.
Although these methods have achieved commendable results, they frequently encounter a trade-off between adherence to instructions and semantic consistency.
Unlike the standard practice of adopting image tokens alongside text tokens as input to VLMs, we provide fine-grained visual features to ground textual instructions, thereby enhancing the reasoning capabilities of the VLM. Our insight is that directly utilizing  LLMs to interpret editing intentions is not sufficiently effective. Consequently, we decouple the process of understanding instructions into two distinct tasks: locating the region of interest and associating the relevant text. We observe that the perceptual skills required for localization are not the strong suit of LLMs. By integrating fine-grained region features into the tokenization process of image features, we can complement the capabilities of LLMs. Specifically, we employ an open-vocabulary object detector~\cite{zhou2022detecting} to identify potential regions of interest and tokenize each region into region tokens, which serve as a supplement to standard image tokens. The enhanced visual tokens, along with instruction tokens, form a new input for the LLM, which implicitly locates the editing regions and articulates the modification intentions through the LLM's powerful reasoning capabilities. 
Introducing region tokens improves the LLM’s ability to perceive the visual world.
 
Once the editing representation is obtained from the LLM response, a direct approach is to employ it for interaction with the diffusion model.
However, due to the semantic gap between the output of VLM and the diffusion model, this approach may result in insufficient guidance. To address this issue, we introduce a Time-Aware Target Injection (TATI) module. As previously noted~\cite{zhang2023prospect,hu2024ella}, the diffusion model emphasizes low-frequency structures (e.g., edges and layouts) in the early stages of the denoising process, while it focuses on high-frequency details (e.g., textures and colors) in the later stages. Therefore, we propose TATI to associate the desired editing region and object shape at the low-frequency semantic level. Subsequently, during the later stages of denoising, TATI captures high-frequency details, linking the attributes of the target. 
To maintain spatial detail consistency between the source image and the edited image, we propose the Hybrid Visual Cross Attention (HVCA) module. HVCA integrates multi-scale visual features to deliver rich visual details, which may play a crucial role in preserving the content of non-edited areas during the intermediate feature evolution process in diffusion models. We utilize learnable queries to filter out the detailed features that need to be preserved through interaction with the text features. These two modules complement each other and enhance the controllability of the denoising process.

By integrating a VLM with fine-grained visual perception and a time-dependent diffusion model, FireEdit achieves high levels of semantic consistency and editability. Extensive quantitative and qualitative results demonstrate that our proposed method outperforms other instruction-based editing techniques in complex scenarios, semantic consistency, and fine-grained editing.
Our main contributions can be summarized as follows:
\begin{itemize}
 \item We propose FireEdit, a novel, versatile, and precise instruction-based image editing method that leverages a region-aware VLM to comprehend editing instructions in complex scenarios.
 \item We propose a time-aware target injection module and a hybrid visual cross-attention module to adaptively control the editing targets and preserve high-frequency details in non-edited regions.
\item Extensive experiments demonstrate that our method outperforms other instruction-based editing methods, establishing a new state-of-the-art in this field.
\end{itemize}

\section{Related Work}
\begin{figure*}[t]
\centering
\setlength{\abovecaptionskip}{5pt}
\includegraphics[width=1\linewidth]{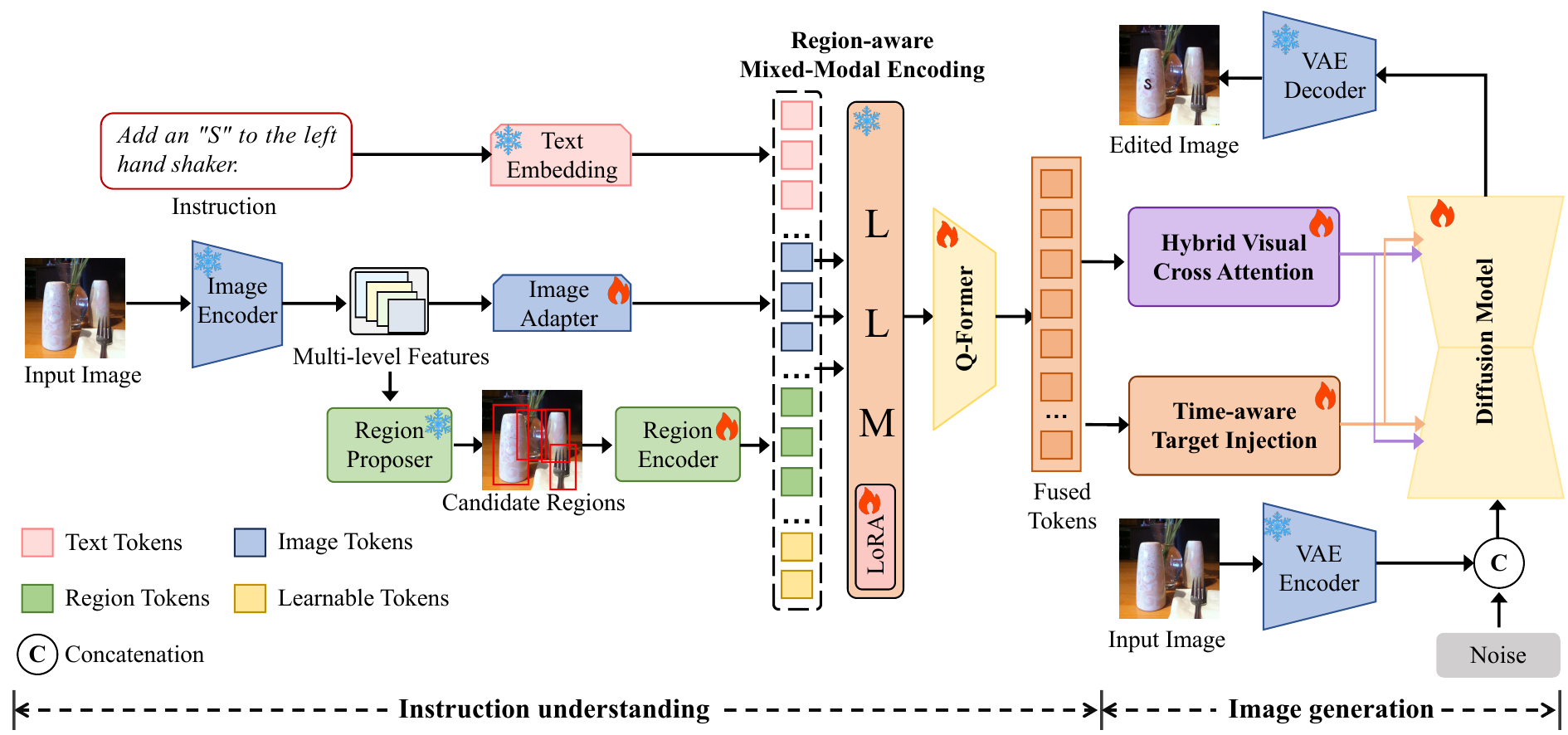}
\caption{The overall framework of FireEdit. The core of FireEdit is to conduct region-aware fusion of multi-modal tokens to promote VLMs and facilitate fine-grained, localized alignments between editing instructions and images. It also introduces a hybrid visual cross-attention module to better preserve image details and a time-aware target injection module to edit targets adaptively.}
\label{fig:model}
\vspace{-10pt}
\end{figure*}
\label{sec:rel work}
\subsection{Instruction-based Image Editing}
The instruction-based image editing task involves using human natural language to modify specified content in the input image while minimizing the distortion of undesired areas. Compared to the text-to-image generative task, the editing task typically entails targeted alterations to specific entities or characteristics within the image. Recently, there are some excellent works~\cite{avrahami2022blended, brooks2023instructpix2pix, zhang2024magicbrush, couairon2023diffedit, li2024zone, hertzprompt,tumanyan2023plug,li2024brushedit} on image editing based on diffusion model. For instance, Prompt-to-Prompt~\cite{hertzprompt} adjust the cross-attention features according to the difference between the source caption and the target caption. Building on this approach, IP2P~\cite{brooks2023instructpix2pix} fine-tuned stable diffusion on a large-scale dataset of source-target-instruction triplets constructed by Prompt-to-Prompt~\cite{hertzprompt}, enabling the model to perform editing under the guidance of simple natural instructions.
Subsequently, MagicBrush~\cite{zhang2024magicbrush} further improves the model's editing capabilities with a real-world dataset. HQ-Edit~\cite{hui2024hq} utilize DALL-E3 and GPT-4V~\cite{openai2023gpt4V} to construct a high-quality synthetic dataset for fine-tuning IP2P. InstructDiffusion~\cite{geng2024instructdiffusion} extends this instruction editing paradigm, trains a general model capable of handling various visual tasks across multiple datasets. UltraEdit~\cite{zhao2024ultraedit} scales up data by combining SDXL with advanced inpainting methods~\cite{nichol2021glide,avrahami2022blended} and trains on a more powerful generative diffusion model~\cite{esser2024scaling}.
To enhance the editing capabilities for complex instructions, SmartEdit~\cite{huang2024smartedit} and MGIE~\cite{fuguiding} introduce VLMs to assist in training diffusion models, facilitate understanding and reasoning for complex instructions. Our method aims to achieve precise localization of editing targets in complex scenarios while preserves high-frequency details in non-edited regions.

\subsection{VLMs with Diffusion Models}

Large language models (LLMs) have demonstrated impressive generalization capability in text generation~\cite{chiang2023vicuna,touvron2023llama}. Providing visual inputs to assist text generation has shown outstanding performances in a range of downstream language-to-vision tasks~\cite{xu2023bridging,li2023blip,liu2024visual,xue2024xgen,rasheed2024glamm,ma2025groma,huang2025densely}. Bridging the gap between LLMs and visual inputs is a key focus of the latest Vision-Language Models (VLMs). 
LLaVA~\cite{liu2024visual} fine-tune LLMs on an instruction-following dataset constructed by GPT-4. xGen-MM~\cite{xue2024xgen} replaces the Q-Former in BLIP-2 with a scalable visual token sampler, trains LLMs on large-scale, high-quality, and diverse datasets to advance various vision-language tasks. Leveraging the strong reasoning capabilities of vision-language models, researchers have begun to introduce them into the field of image generation. GILL~\cite{koh2024generating} is a pioneering work in this attempt, guides diffusion models to generate images consistent with text by using text representations inferred from vision-language models. MGIE~\cite{fuguiding} extends this to image editing, leveraging VLMs to learn concise expressive instructions. SmartEdit~\cite{huang2024smartedit} introduces a bidirectional interactive module to facilitate mutual understanding between the text representations outputted by VLMs and the input image features, addressing the issue of CLIP's text encoder's insufficient understanding of complex instructions. ILLUM~\cite{wang2024illume} establishes a unified VLM with the capability of understanding images and charts, as well as image generation and editing.
Unlike previous methods that train a custom VLM, PromptFix~\cite{yu2024promptfix} utilizes the extensive world knowledge and understanding capabilities of the existing LLaVA to enhance text prompts.
Our approach also focuses on training a dedicated VLM for editing tasks. Our insight is that a general VLM often struggles to effectively understand overly simple or complex instructions, necessitating the decoupling of the instruction understanding process into locating and associating target objects. We integrate fine-grained region tokens into standard image tokenization, providing the LLM with more visual imagination possibilities, thereby obtaining more robust and accurate text representations of user editing intentions.
\section{Preliminary}
The instruction-based editing method aims to modify a specific area of a given image $x$ according to the provided natural language instruction $c_T$ to obtain the edited image. InstructPix2Pix (IP2P), a latent diffusion model, is a pioneering work in this field. IP2P fine-tunes the pre-trained Stable Diffusion (SD) model on a synthesized large-scale dataset consisting of input-target-instruction triplets. Given a target image $y$ and an image encoder $\mathcal{E}$, the diffusion process adds noise to the encoding latent $z=\mathcal{E}(y)$, resulting in a noisy latent $z_t$, with the noise level increasing over timestep $t\in T$. A diffusion model $\epsilon_{\delta}$, given the image condition $c_I$ and the text instruction condition $c_T$, is trained to minimize
\begin{equation}
\mathcal{L}_{diff} =\mathbb{E}_{\mathcal{E}(y),\mathcal{E}(c_I),c_T,\epsilon,t}||\epsilon - \epsilon_{\delta}(z_t,t,\mathcal{E}(c_I),c_T)||_2^2,
\end{equation}
where $\epsilon$ is the noise added to the noisy latent $z_t$.
IP2P extends the classifier-free guidance (CFG) strategy to obtain score estimates in dual condition pattern and introduces an image guidance scale $s_I$ and an instruction guidance scale $s_T$ to balance the guidance strengths from both conditions: 
\begin{equation}
    \begin{split}
    \tilde{\epsilon}_{\delta}(z_t,t,c_I,c_T) &= \epsilon_\delta(z_t,t,\emptyset,\emptyset) \\
     &+ s_I\cdot(\epsilon_\delta (z_t,t,c_I,\emptyset)-\epsilon_\delta(z_t,\emptyset,\emptyset)
     \\
     & + s_T\cdot(\epsilon_\delta(z_t,t,c_I,c_T)-\epsilon_\delta(z_t,t,c_I,\emptyset)). 
    \end{split}
    \label{eq1}
\end{equation}
During inference, our method also derives score estimates in this manner. An increase in $s_I$ retains semantic details but weakens the instructions, while an increase in $s_T$ strengthens the instructions, potentially leading to over-editing.
\section{Methodology}
\label{sec: method}
In this section, we provide a detailed introduction to the proposed FireEdit framework. FireEdit is designed to achieve precise and high-quality image editing tasks based on text instructions. Unlike existing methods that mainly rely on text encoders to understand user intentions, we propose to fine-tune a VLM by region-aware fusion of multi-modal tokens, so that we can exploit the extensive world knowledge of VLM and achieve model fine-grained alignments between images and instructions. As shown in Figure \ref{fig:model}, the core components of FireEdit include a region-aware mixed-model encoder for token fusion  (Sec. \ref{sec:encoder}), a hybrid visual cross-attention module for preserving image details (Sec. \ref{sec:HVCA}), and a time-aware target injection module for incorporating target information into images adaptively  (Sec. \ref{sec:TTI}).

\subsection{Region-aware Mixed-Modal Encoding}
\label{sec:encoder}
Given an input image $x$, a text instruction $c$, which is tokenized into a sequence $(s_1,s_2,\dots, s_T)$, our goal is to achieve precise modification to $x$ under the guidance of $c$ without impacting other undesired regions. As shown in Figure \ref{fig:model}, our model consists of a VLM and a diffusion model. Unlike the previous method that only uses the visual features extracted by the image encoder $\mu_{\nu}(\cdot)$ along with the text embeddings from $(s_1,s_2,\dots, s_T)$ as the input of the LLM, $\mathcal{V}(\cdot;\theta)$, our method incorporates fine-grained visual tokens. Specifically, we obtain multi-level features from $\mu_{\nu}(\cdot)$ and use a sophisticated detector $\mathcal{D}(\cdot)$ to obtain potential candidate regions. A region encoder $\mathcal{R}(\cdot)$ is deployed to extract features of these regions of interest, which will serve as region tokens to provide fine-grained visual information for LLM. We define the process for obtaining the embeddings of region tokens as follows:
\begin{equation}
    \mathcal{P} = \mathcal{R}(\mathcal{D}(x)),
\end{equation}
here $\mathcal{P}\in\mathbb{R}^{L_r\times D}$, $L_r$ indicates the length of the region tokens and $D$ is the dimension of the embeddings.

The response of VLM is represented as a sequence of discrete tokens, which cannot be directly utilized by diffusion models. Current methods \cite{koh2024generating,huang2024smartedit} address this by expanding the vocabulary of LLaVA \cite{liu2024visual} with special [IMG] tokens. Similarly, our method introduces $r$ additional special tokens, denoted as $\mathcal{Q}=\{[\text{IMG}_1], \dots, [\text{IMG}_r]\}$. Based on this, we incorporate a learnable embedding matrix $E$ into the LLM's embeddings. The learning objective of the VLM is thus transformed to predict these $r$ tokens and extract their corresponding latent states as editing representations. 
Therefore, we restructure the input of the VLM to include a cascade of holistic image feature tokens $\mu_{\nu}(x)$, region tokens $\mathcal{P}$, text embeddings $c$, and learnable query tokens $\mathcal{Q}$. The process of understanding the input instructions can be formalized as follows:
\begin{equation}
    \begin{split}
    \mathcal{H} &= [c,\mu_{\nu}(x),\mathcal{P},\mathcal{Q}], \\
    e &=\mathcal{V}(\mathcal{H};\theta),
    \end{split}
\end{equation}
$[\cdot, \cdot]$ refers to the concatenation operation, and $e \in \mathcal{R}^{r \times D}$ corresponds to the hidden embeddings of $r$ special tokens.
\begin{figure*}[tp]
	\centering
\setlength{\abovecaptionskip}{5pt}
\includegraphics[width=0.95\linewidth]{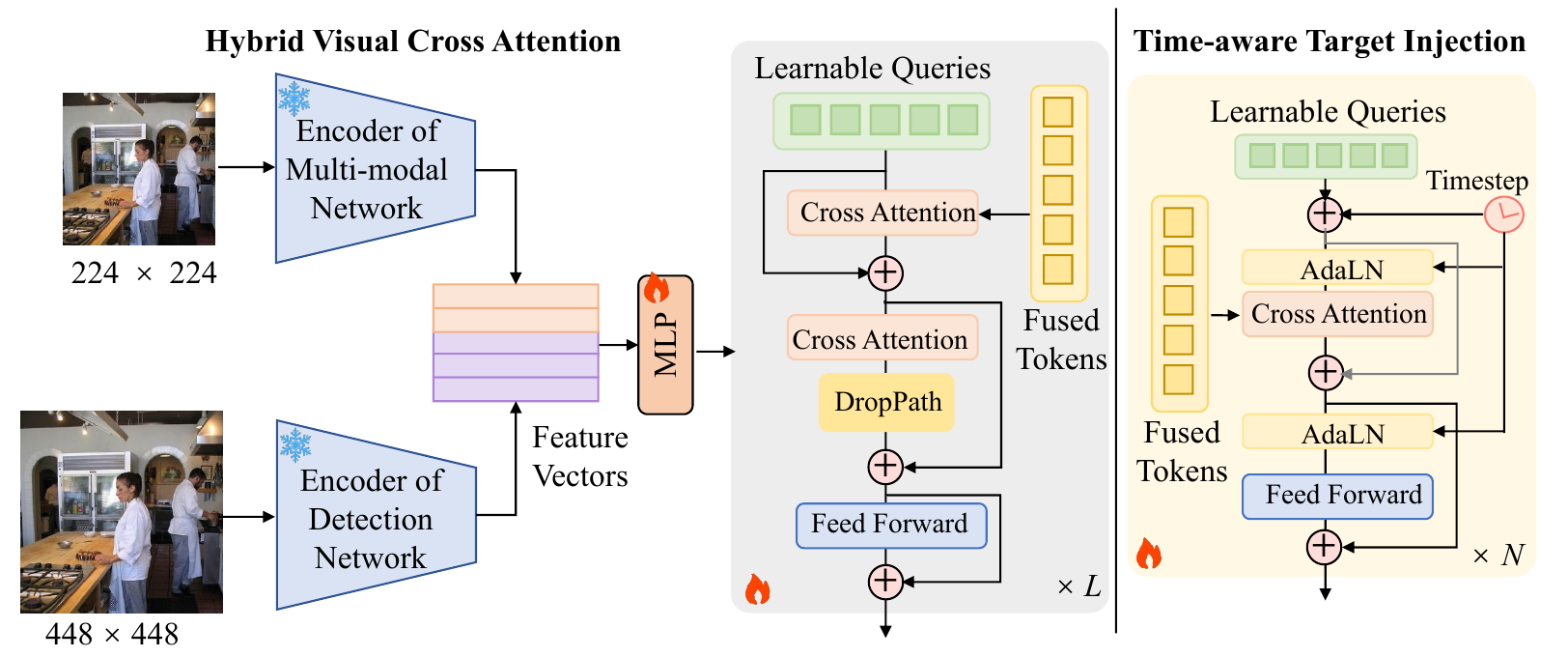}
	\caption{The proposed hybrid visual cross attention module (HVCA, left) and the time-aware target injection module (TATI, right). 
    HVCA exploits hybrid encoders of pre-trained networks to balance between global semantic information (e.g., multi-modal alignment networks like CLIP) and local details. TATI utilizes timestep embeddings to incorporate target information into the denoising process.}
 \label{fig:module}
 \vspace{-15pt}
\end{figure*}
Due to the semantic gap between the text embedding space of the diffusion model and the output of the LLM, an additional Q-Former module is required to bridge them. Finally, we obtain the edited representation $e_t$, which is derived from $e$ through the Q-Former, is used to guide the diffusion model in the subsequent steps.

To effectively train the VLM, we freeze most of the parameters of the LLM and employ LoRA \cite{hu2021lora} for fine-tuning. Additionally, we freeze the image encoder and train an adapter $W$ to adapt the image features to the latent space of the LLM. We use a pre-trained Deformable DETR \cite{zhu2020deformable} to extract region proposals and a region encoder transforms the region features into region tokens. Consequently, the $i$-th [$\text{IMG}_i$] token is generated by minimizing the negative log-likelihood based on the previously generated tokens
\begin{equation}
   \begin{split}
       \mathcal{L}_{VLM} = - \sum_{i=1}^{r}\log p(\theta\cup\Delta\theta)([\text{IMG}_i]|\mu_{\nu}(x)^TW,\\
       \mathcal{P}, c, [\text{IMG}_1],\dots,[\text{IMG}_{i-1}]),
   \end{split}
\end{equation}
where $\Delta\theta$ denotes the parameter of LoRA.

\subsection{Hybrid Visual Cross Attention}
\label{sec:HVCA}
To prevent the loss of viusal details in non-target areas, we introduce a Hybrid Visual Cross Attention (HVCA) module. As shown in Figure \ref{fig:module}, we cascade the visual features extracted by the CLIP image encoder \cite{radford2021learning} and DINOv2 \cite{oquab2023dinov2} as visual inputs. The use of different image encoders is intended to capture rich visual details at various pixel levels. These features are then enhanced through interaction with text-related features to obtain reinforced visual features. Our HVCA consists of $L$ attention blocks, each comprising two cross-attention layers and a feed-forward network. The IP-Adapter \cite{ye2023ip} directly injects features extracted by the image encoder into the denoising network, which makes it challenging to maintain high-frequency details, especially in the field of image editing. By introducing visual content with context-aware capabilities, we ensure that the generated images have higher fidelity while maintaining editability.

Following the IP-Adapter \cite{ye2023ip}, we decouple the cross-attention layers to inject the refined visual embeddings $v$ and the edited representation $e_t$ into the cross-attention layers of the denoising network. Specifically, given the query features $Z$, the edited representation $e_t$, and the visual embeddings $v$, the output $Z^{\prime}$ of the cross-attention can be defined as follows:
\begin{equation}
    Z^{\prime} = \text{Softmax}(\frac{QK_1^T}{\sqrt{d}})V_1 + \lambda\cdot \text{Softmax}(\frac{QK_2^T}{\sqrt{d}})V_2,
\end{equation}
where $Q=ZW_q$, $K_1=e_tW_{k_1}$, $V_1=e_tW_{v_1}$, and $K_2=vW_{k_2}$, $V_2=vW_{v_2}$ are query, key, and value matrices. $\lambda$ is weight factor, and $d$ represents the feature dimension. 

\subsection{Time-aware Target Injection}
\label{sec:TTI}

Previous research has demonstrated that diffusion models for text-to-image generation tend to focus on low-frequency structures during the early stages of denoising, while concentrating on high-frequency details in the later stages. Instruction-based image editing retains this characteristic, attending to different transformations at various stages of the denoising process \cite{zhao2024instructbrush}. Unlike \cite{zhao2024instructbrush}, which optimizes instruction features by splitting different denoising stages, we associate sampling timesteps for editing instructions using a resampler that incorporates time embeddings. We refer to the proposed module as the Time-Aware Target Injection (TATI) module, which is inspired by \cite{hu2024ella}. Although structurally similar, the motivations are distinct. TATI incorporates timestep information across the diffusion process, optimizing text representation utilization. Specifically, our TATI consists of $N$ stacked resamplers \cite{alayrac2022flamingo}, where temporal information is integrated into adaptive Layer Normalization \cite{peebles2023scalable} within each resampler to establish time dependencies.

\subsection{Optimization Objectives}
During the diffusion process, we follow the design principles of IP2P \cite{brooks2023instructpix2pix}. We concatenate the encoded image latent $\mathcal{E}(x)$ and the noisy latent $z_t$ as input to the denoising network $\epsilon_{\delta}$. The difference lies in that we substitute the text embeddings obtained by the CLIP text encoder with $e_t$, and use it along with the visual feature $v$ through the decoupled cross-attention layers of the UNet to participate in the denoising process. The optimization process for $\epsilon_{\delta}$ can be expressed as follows:
\begin{equation}
\mathcal{L}_{diff} =\mathbb{E}_{\mathcal{E}(y),\mathcal{E}(x),c,\epsilon,t}||\epsilon - \epsilon_{\delta}(t,\mathcal{C}(z_t,\mathcal{E}(x)),e_t,v)||_2^2,
\end{equation}
where $\epsilon\sim\mathcal{U}(0,1)$, $t$ is the sampling time, $y$ is the target edited images and $\mathcal{C}$ denotes the operation of concatenation along the channel dimension.
The combined total loss function for the VLM and the diffusion model is defined as
\begin{equation}
    \mathcal{L}_{total} = \mathcal{L}_{VLM} + \mathcal{L}_{diff}.
\end{equation}

\section{Experiments}
\begin{figure*}[t]
	\centering
    \setlength{\abovecaptionskip}{5pt}
  \includegraphics[width=1\linewidth]{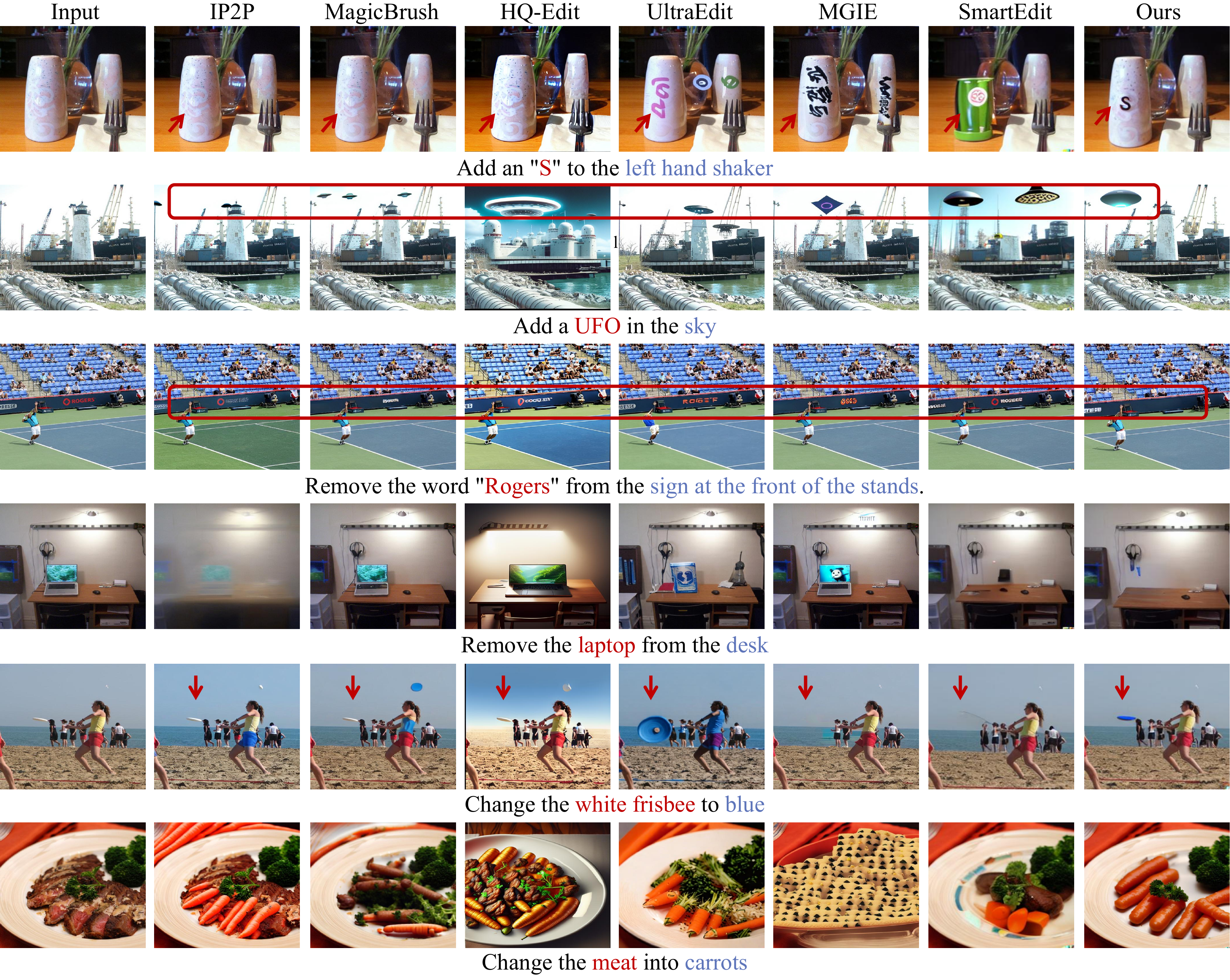}
	\caption{Qualitative comparison. We compare the editing performance of FireEdit with SOTA methods on the Emu Edit test set. Each editing instruction is written below each row of images. Compared with other SOTA methods, our approach is superior in accurately locating the edited objects or regions and preserving the detailed information of the input image.}
  \vspace{-5pt}
 \label{fig:qua}
\end{figure*}
\begin{table}[tp]
    \caption{Results on the Emu Edit test. We compare our approach with other instruction-based image editing baselines. 
    $ \uparrow$ indicates the higher the better, while $ \downarrow$ indicates the lower the better. 
    The best results are in bold.}
   
    \label{tab:emu}
    \centering
    \setlength{\belowcaptionskip}{-10pt}
    \resizebox{\linewidth}{!}{
    \begin{tabular}{ccccccc}
        \toprule[1pt]
        \multicolumn{1}{c}{Method} &
        \multicolumn{1}{c}{L1$\downarrow$} &
        \multicolumn{1}{c}{CLIP-I$\uparrow$} &
        \multicolumn{1}{c}{DINO$\uparrow$} &
         \multicolumn{1}{c}{LPIPS$\downarrow$} &
        \multicolumn{1}{c}{CLIP-T$\uparrow$} &
        \multicolumn{1}{c}{User Study$\uparrow$} \\
        \midrule[1pt]
         IP2P & 0.1225& 0.8570  &0.7616 & 0.2723 &0.2715  & 1.83\%      \\
        MagicBrush& 0.0991&  0.8670& 0.7810 & 0.2551  &0.2744  & 5.00\%     \\
        HQ-Edit & 0.2541 & 0.7095& 0.5404 & 0.5397 &0.2566  &4.92\%     \\
        UltraEdit& 0.0838& 0.8120 & 0.7406 & 0.2826 &0.2773 &5.00\%  \\
        MGIE& 0.1628 & 0.7456 & 0.5944 & 0.3801 & 0.2314 & 1.25\%      \\
        SmartEdit & 0.1186 & 0.8592& 0.7705 & 0.2717  &0.2740    & 3.33\%   \\
        \midrule[1pt]
        Ours  & \textbf{0.0574}    & \textbf{0.9140}   & \textbf{0.8829} &  \textbf{0.1373}  &\textbf{0.2783}  &\textbf{78.67\%}
        \\ \bottomrule[1pt]
   \end{tabular}
   }
   \vspace{-15pt}
\end{table}
\subsection{Implementation Details}
\textbf{Model Architecture:} Following \cite{huang2024smartedit,fuguiding}, our visual large language model retains the large language model component of LLAVA-7B, with the visual encoder being DINOv2 and a two-layer linear layer connecting the LLM and the image encoder. We use Deformable DETR \cite{zhu2020deformable} as the region detector and ROIAlign \cite{rasheed2024glamm} as the region encoder, and their pre-trained weights are derived from \cite{ma2025groma}. Our Q-Former uses the same structure as \cite{huang2024smartedit}. TATI contains 2 units and 77 learnable queries, while the HVCA module has 2 units and 16 learnable queries. For LLM, we freeze its parameters and use LoRA for fine-tuning, with rank and alpha of 8 and 16, respectively. We implement experiments on Stable Diffusion 1.5, and its parameters are initialized by IP2P \cite{brooks2023instructpix2pix}.

\noindent\textbf{Training Dataset:} Our training data can be divided into three categories: (1) segmentation dataset, which comes from COCOStuff \cite{caesar2018coco}, RefCOCO \cite{yu2016modeling}, GRefCOCO \cite{liu2023gres}; (2) image editing dataset, including InstructPix2Pix \cite{brooks2023instructpix2pix}, MagicBrush \cite{zhang2024magicbrush}, Ultraedit \cite{zhao2024ultraedit}, and ReasonEdit \cite{huang2024smartedit}; (3) visual question answering dataset, LLAVA-instruct-150k \cite{liu2024visual}.

\noindent\textbf{Benchmark and Metrics:}
We validate our approach on two popular image editing test sets: MagicBrush test \cite{zhang2024magicbrush} and Emu Edit test \cite{sheynin2024emu}. MagicBrush evaluates the model across multiple metrics by comparing the generated edited images to the ground truth images and the corresponding captions. Following \cite{zhang2024magicbrush,zhao2024ultraedit}, we use L1 distance, L2 distance, CLIP image similarity (CLIP-I), and DINO similarity as metrics. The Emu Edit test set verifies the edited images against the source images and the target captions. In addition to the above metrics, we also use CLIP text-image similarity and LPIPS. 

\noindent\textbf{Baseline Models:} We make comparisons with the state-of-the-art instruction-based image editing methoss, including IP2P \cite{brooks2023instructpix2pix}, MagicBrush \cite{zhang2024magicbrush}, HIVE \cite{zhang2024hive}, HQ-Edit \cite{hui2024hq}, UltraEdit \cite{zhao2024ultraedit}, MGIE \cite{fuguiding},and SmartEdit \cite{huang2024smartedit}. In addition, we compare with description-based image editing methods, including SDEdit \cite{mengsdedit}, NTI \cite{mokady2023null}, GLIDE \cite{nichol2021glide}, and BleDiff \cite{avrahami2022blended} on the MagicBrush test set.
\begin{table}[t]
    \caption{Results on the MagicBrush test set. We include both single-turn and multi-turn settings. 
    The results demonstrate that our method outperforms other VLM-guided editing techniques. 
    }
    \setlength{\belowcaptionskip}{-5pt} 
    \label{tab:mb_test}
    \centering
    \resizebox{1.0\linewidth}{!}
    {
        \begin{tabular}{cclcccc}
            \toprule
            \multicolumn{2}{c}{Settings} &
            \multicolumn{1}{c}{Method}&
            \multicolumn{1}{c}{L1$\downarrow$} &
            \multicolumn{1}{c}{L2$\downarrow$} &
            \multicolumn{1}{c}{CLIP-I$\uparrow$} &
            \multicolumn{1}{c}{DINO$\uparrow$} 
            \\ 
            \midrule
            \multicolumn{2}{c}{\multirow{12}{*}{Single-turn}}& \multicolumn{5}{c}{Global Description-guided} \\ 
            \cline{3-7}
            &\multicolumn{1}{c}{}& SDEdit& 0.1014& 0.0278& 0.8526& 0.7726 \\
            &\multicolumn{1}{c}{}& NTI & 0.0749& 0.0197& 0.8827& 0.8206 \\
            &\multicolumn{1}{c}{}& GLIDE & 3.4973& 115.8347& 0.9487& 0.9206 \\
            &\multicolumn{1}{c}{}& BleDiff & 3.5631& 119.2813& 0.9291& 0.9644 \\ \cline{3-7}
            &\multicolumn{1}{c}{}&\multicolumn{5}{c}{Instruction-guided} \\ \cline{3-7}
            &\multicolumn{1}{c}{}& HIVE& 0.1092& 0.0380& 0.8519& 0.7500 \\
            &\multicolumn{1}{c}{}& IP2P& 0.1141& 0.0371& 0.8512& 0.7437 \\
            &\multicolumn{1}{c}{}& MagicBrush& 0.0625& 0.0203& 0.9332& 0.8987 \\
            \cline{3-7}
            &\multicolumn{1}{c}{}& MGIE& 0.1624& 0.0669& 0.7454& 0.5771 \\
            &\multicolumn{1}{c}{}& SmartEdit& 0.0936& 0.0377& 0.8945& 0.8201 \\
            &\multicolumn{1}{c}{}& Ours&\textbf{0.0701} &\textbf{0.0238} &\textbf{0.9131} & \textbf{0.8619}
            \\ 
            \midrule
            \multicolumn{2}{c}{\multirow{12}{*}{Multi-turn}}& \multicolumn{5}{c}{Global Description-guided} \\ \cline{3-7}
            &\multicolumn{1}{c}{}& SDEdit& 0.1616& 0.0602& 0.7933& 0.6212 \\
            &\multicolumn{1}{c}{}& NTI& 0.1057& 0.0335& 0.8468& 0.7529 \\
            &\multicolumn{1}{c}{}& GLIDE & 11.7487& 1079.5997& 0.9094& 0.8498 \\
            &\multicolumn{1}{c}{}& BleDiff & 14.5439& 1510.2271& 0.8782& 0.7690 \\ \cline{3-7}
            &\multicolumn{1}{c}{}&\multicolumn{5}{c}{Instruction-guided} \\ \cline{3-7}
            &\multicolumn{1}{c}{}& HIVE& 0.1521& 0.0557& 0.8004& 0.6463 \\
            &\multicolumn{1}{c}{}& IP2P& 0.1345& 0.0460& 0.8304& 0.7018 \\
            &\multicolumn{1}{c}{}& MagicBrush& 0.0964& 0.0353& 0.8924& 0.8273 \\
            \cline{3-7}
            &\multicolumn{1}{c}{}& MGIE & 0.1912& 0.0836& 0.7008& 0.5208 \\
            &\multicolumn{1}{c}{}& SmartEdit & 0.1218& 0.0510& 0.8601& 0.7538 \\
            &\multicolumn{1}{c}{}& Ours&\textbf{0.0911} &\textbf{0.0326} &\textbf{0.8819} & \textbf{0.8010} \\
            \bottomrule
        \end{tabular}
        }
        \vspace{-15pt}
\end{table}

\subsection{Comparisons with SOTA models}
\noindent\textbf{Quantitative Evaluation.} We conducted quantitative evaluations on the Emu Edit and MagicBrush test sets. Table \ref{tab:emu} shows the quantitative results on the Emu Edit test set, revealing that our method outperforms competitors in metrics related to semantic structure and quality, suggesting that FireEdit more effectively preserves non-targeted editing regions.
In addition, compared with VLM-based editing methods such as SmartEdit and MGIE, our approach significantly enhances semantic consistency without compromising its robust editing capabilities.

Table \ref{tab:mb_test} shows the quantitative results on the MagicBrush benchmark, where we conducted a comprehensive comparison across both single-turn and multi-turn editing scenarios. Our method demonstrates a distinct advantage over VLM-guided counterparts in preserving non-editing areas. Notably, MagicBrush performs exceptionally well across various metrics, likely due to biases introduced from its training set. 
To further demonstrate the effectiveness of our approach, we conduct user studies from three perspectives: editing instruction alignment, semantic consistency, and image quality. Table \ref{tab:emu} shows the user preference for our approach, with higher user scores. More details of the User Study can be found in the supplementary materials.

\noindent\textbf{Qualitative Comparsion.} 
In Figure \ref{fig:qua}, we illustrate the editing results of the SOTA method and our method. We selected 6 sets of edited images across 3 types of editing actions (add, remove, change), and the inputs were all from real images. Our method achieves precise localization of the edited area while preserving the remaining pixels, which is particularly important for perceiving high-frequency details (texture, color, etc.). For example, in the first row of Figure \ref{fig:qua}, our method can accurately add the letter "S" while recognizing the location of the target pointed to by "the left hand shaker" in the input image. In contrast, other methods either fail to perceive the correct location or do not accurately add the "S" symbol. In the second row, HQ-Edit correctly adds the UFO, but causes the degradation of the semantic structure, similar issues are observed with UltraEdit and SmartEdit. Our method, however, achieves local editing without changing the semantic structure. In the fifth row, our method can locate the "white Frisbee" from the complex scene and modify it to blue. IP2P fails to locate the desired editing object, and UltraEdit modifies irrelevant objects (the color of the lady's top next to the Frisbee is changed).
\begin{figure}[t]
\centering
\setlength{\abovecaptionskip}{5pt}
\includegraphics[width=\linewidth]{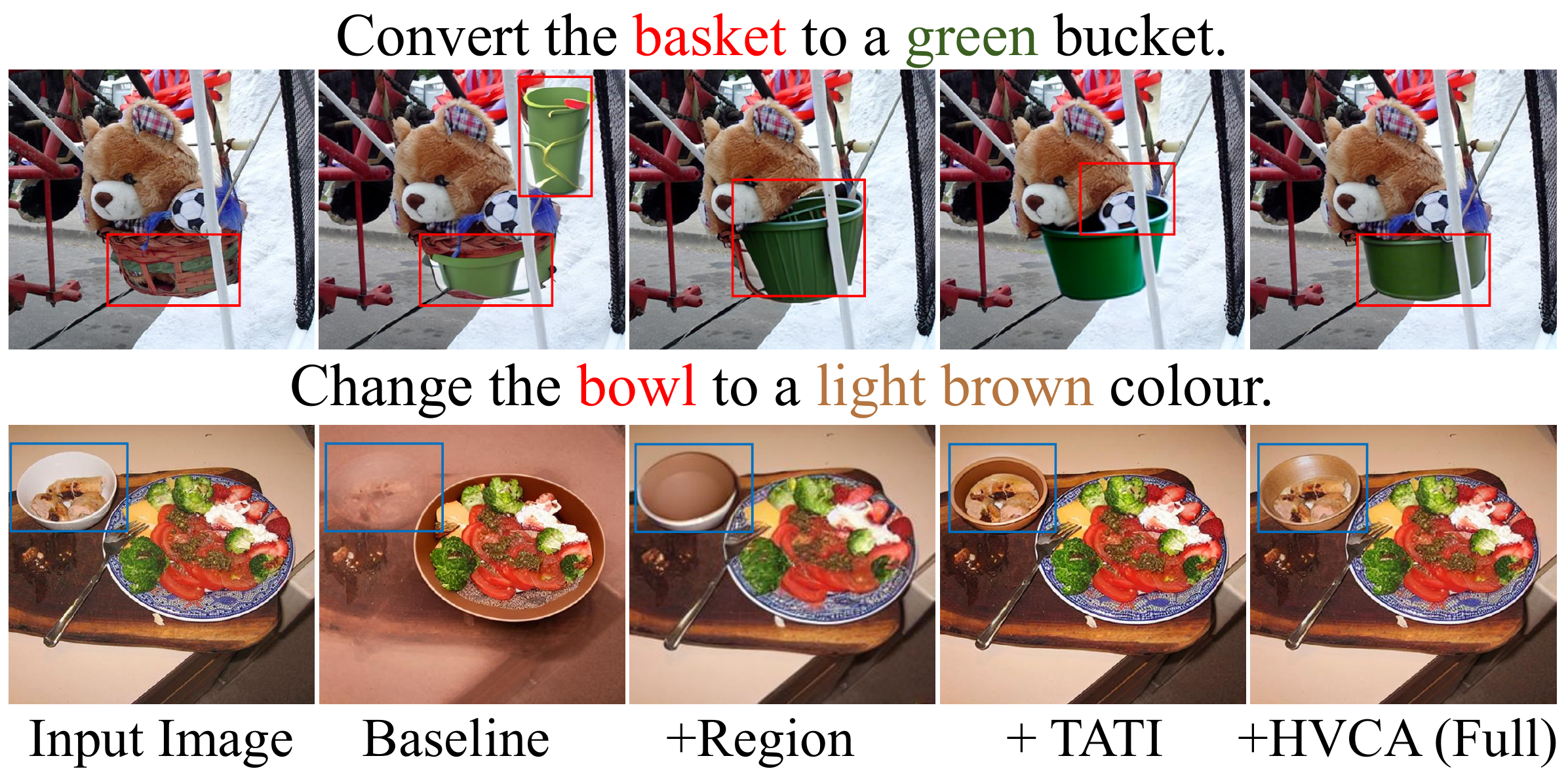}
\caption{Ablation studies for components in our method.}
\label{fig:abl}
\vspace{-15pt}
\end{figure}
\subsection{Ablation Study}
The quantitative results of the ablation study we conducted are presented in Table \ref{tab:ab}, and some representative visualizations can be seen in Figure \ref{fig:abl}.

\noindent\textbf{Effect of Region-aware Mixed-Modal Encoding (Region).}
To validate the effectiveness of region-aware mixed-modal encoding, we remove this component and make comparison on the Emu edit test set. As illustrated in Figure \ref{fig:abl}, the model struggles to accurately identify the editing targets without the region tokens, leading to the unintended alteration of content in undesired areas. This is supported by the metrics assessing semantic structure presented in Table \ref{tab:ab}. The incorporation of region tokens significantly improves the VLM's capability to locate the intended targets.

\noindent\textbf{Effect of Time-aware Target Injection (TATI).}
We confirm the crucial role of the TATI module in balancing semantic consistency and editability.
As shown in Table~\ref{tab:ab}, we observe a significant improvement in the metrics measuring semantic consistency, while the instruction-following metrics remained at a satisfactory level. The TATI module facilitates adaptive adjustments to the guidance strength at different denoising stages. As shown in the first row of Figure \ref{fig:abl}, TATI mitigates the erosion of non-editing regions. 
The model's enhanced instruction-following capability allows for precise modifications to the editing targets while preserving the content of non-editing regions.

\noindent\textbf{Effect of and Hybrid Visual Cross Attention (HVCA).}
To demonstrate the effectiveness of HVCA, we integrate it with region tokens and the TATI module.
As shown in Table \ref{tab:ab}, after implementing HVCA, our method demonstrates improvements in the L1, CLIP-T, and LPIPS metrics, indicating its effectiveness in retaining semantic details. As illustrated in Figure \ref{tab:ab}, HVCA integrates visual features across multiple scales, thereby enhancing visual detail. The interaction between the fused visual features and the editing instructions strengthens the preservation of high-frequency details, effectively incorporating visual content into the denoising process of the diffusion model.
\begin{table}[t]
    \caption{Ablation study on the Emu Edit test. The model's performance benefits from all of the components.}
    \label{tab:ab}
    \centering
    \setlength{\belowcaptionskip}{0pt} 
    \resizebox{1.0\linewidth}{!}
    {
    \begin{tabular}{ccccccccc}
        \toprule[1pt]
        \multicolumn{1}{c}{Region} &
        \multicolumn{1}{c}{TATI} &
        \multicolumn{1}{c}{HVCA} &
        \multicolumn{1}{c}{L1$\downarrow$} &
        \multicolumn{1}{c}{CLIP-I$\uparrow$} &
        \multicolumn{1}{c}{LPIPS$\downarrow$} &
        \multicolumn{1}{c}{CLIP-T$\uparrow$} & 
        \\
        \midrule[1pt]
         -& -& -& 0.0897& 0.8501& 0.3174  &0.2770    \\
         \checkmark& -&  -& 0.0859 & 0.8709 & 0.2391& 0.2762    \\
        \checkmark&\checkmark &  - & 0.0609 &  \textbf{0.9258}& 0.1580 & 0.2746     \\
        \checkmark&\checkmark & \checkmark& \textbf{0.0574} & 0.914 & \textbf{0.1373}  & \textbf{0.2783}
        \\ \bottomrule[1pt]
   \end{tabular}
   }
    \vspace{-10pt}
\end{table}

\section{Limitations}
Although FireEdit outperforms other baseline methods in comprehending simple text instructions within complex scenarios, it does not support reference images as input. Like other instruction-based image editing techniques, our approach aims to tackle challenging tasks such as altering poses and repositioning objects. In future work, a promising direction is to incorporate reference images to enhance control over the editing target and leverage visual autoregression \cite{tian2024visual} to push the boundaries of image editing.
\section{Conclusion}
In this paper, we propose FireEdit, a novel instruction-based image editing method that achieves precise localization of desired editing regions in complex scenarios through a region-aware vision language model. We introduce a region-aware multi-modal encoder in FireEdit to interpret editing instructions in a fine-grained manner, without the need for manually selected masks or reference regions. Furthermore, FireEdit introduces a hybrid visual cross-attention module and a time-aware target injection module to capture the characteristics of non-target areas and control the editing process adaptively. Extensive experimental results on two public datasets demonstrate that FireEdit outperforms state-of-the-art methods in terms of instruction adherence and semantic consistency.

\section*{Acknowledgements}
This work is supported by National Key Research and Development Program of China (2024YFE0203100), Shenzhen Science and Technology Program No.GJHZ20220913142600001, National Natural Science Foundation of China (NSFC) under Grants No.62476293 and No.62372482, Nansha Key R\&D Program under Grant No.2022ZD014 and General Embodied AI Center of Sun Yat-sen University.

{
    \small
    \bibliographystyle{ieeenat_fullname}
    \bibliography{main}
}

\end{document}